\documentclass{article}
\usepackage{custom_arxiv}

\usepackage{amsmath,amsfonts,bm}

\def\eqref#1{equation~\ref{#1}}

\def\1{\bm{1}}

\DeclareMathAlphabet{\mathsfit}{\encodingdefault}{\sfdefault}{m}{sl}
\SetMathAlphabet{\mathsfit}{bold}{\encodingdefault}{\sfdefault}{bx}{n}

\usepackage{hyperref}
\usepackage{url}
\usepackage{graphicx}
\usepackage{xcolor}
\usepackage{algorithm}
\usepackage{algpseudocode}
\usepackage{subfig}
\def\algbackskip{\hskip-\ALG@thistlm}
\usepackage{paralist}

\title{Approximating DTW with a convolutional neural network on EEG data}
\author{Hugo Lerogeron  \\ Saagie \\ LITIS \\ Rouen \\ \texttt{hugo.lerogeron@saagie.com}
\And Romain Picot-Clémente  \\ Saagie \\ Rouen \\ \texttt{romain@saagie.com}
\And Alain Rakotomamonjy \\ Paris \\ \texttt{alain.rakoto@insa-rouen.fr}
\And Laurent Heutte \\ LITIS \\ Rouen \\ \texttt{laurent.heutte@univ-rouen.fr}
}

\newcommand{\mycomment}[1]{}

\begin{document}
\maketitle
\begin{abstract}
Dynamic Time Wrapping (DTW) is a widely used algorithm for measuring similarities between two time series. It is especially valuable in a wide variety of applications, such as clustering, anomaly detection, classification, or video segmentation, where the time-series have different timescales, are irregularly sampled, or are shifted. However, it is not prone to be considered as a loss function in an end-to-end learning framework because of its non-differentiability and its quadratic temporal complexity. While differentiable variants of DTW have been introduced by the community, they still present some drawbacks: computing the distance is still expensive and this similarity tends to blur some differences in the time-series.
In this paper, we propose a fast and differentiable approximation of DTW by comparing two architectures: the first one for learning an embedding in which the Euclidean distance mimics the DTW, and the second one for directly predicting the DTW output using regression. We build the former by training a siamese neural network to regress the DTW value between two time-series. Depending on the nature of the activation function, this approximation naturally supports differentiation, and it is efficient to compute. We show, in a time-series retrieval context on EEG datasets, that our methods achieve at least the same level of accuracy as other DTW main approximations with higher computational efficiency. We also show that it can be used to learn in an end-to-end setting on long time series by proposing generative models of EEGs.

\end{abstract}
\section{Introduction}
Proposed by Sakoe and al \cite{sakoe1978dynamic},  Dynamic Time Wrapping (DTW) algorithm is an alignment-based similarity measure for temporal sequences. Initially used for speech applications, its properties, notably its invariance to time shifts and its ability to compare series of different lengths, make the DTW useful in various time-series related applications. For instance, Seto and al \cite{seto2015multivariate} make use of the DTW to create meaningful features for human activity recognition, Laperre and al \cite{laperre2020dynamic} employ the DTW as a regularization tool in disturbance storm time forecasting and Zifan and al \cite{zifan2006automated} consider DTW on piecewise approximation of time series to segment ECG data. 
Nevertheless, due to its non-differentiability (see Tavenard \cite{DTW_blog}), DTW  can not be considered as a loss function for end-to-end training of deep neural networks.
To circumvent those limitations, differentiable approximations of the DTW have been proposed, such as SoftDTW by Cuturi and al \cite{softDTW}, which notably replace the min operator by a softmin.

While this approximation enables kernel machine and end-to-end deep neural network training, it keeps the quadratic complexity in time of DTW which creates a running time problem for applications in which longer time series are considered such as EEG signals. For instance, the widely used SleepEDF dataset introduced by Kemp and al \cite{SleepEDF} uses splits of size 3000.
Therefore, in order to be able to use DTW as a loss in an end-to-end training on EEG signals, we propose a neural model that approximates DTW similarity between two time-series. 

To do so, we propose two architectures to compare : an encoder-decoder scheme in which the backbone is a siamese convolutional neural network and a direct regression model. We show that this enables to obtain an accurate, scalable and differentiable approximation of DTW. \bigskip \newline
In this paper, our contributions are the following ones:
\begin{compactitem}    
    \item we compare a direct regression architecture and a siamese encoder-decoder inspired by Courty and al \cite{courty2017learning} to approximate DTW;
    \item we show how such an approximation is fast, more faithful to the objective function than other approximations (namely FastDTW \cite{salvador2007toward} and SoftDTW \cite{softDTW}) and can be used in end-to-end training;
    \item we show how such an approximation can be transferred to others similar EEG signals using another public dataset. 
\end{compactitem}
After considering related works in Section 2, we detail our approach used to approximate DTW on time series in Section 3 and our experimental setup in Section 4. We then discuss in Section 5 the differentiability, time efficiency and performance on classification tasks of our proposed method. We conclude in Section 6 and draw future works from our results.

\section{Related Works}
\mycomment{
The objective of the DTW algorithm is to compute the distance between two time series $ x, x' \in \mathbb{R}^{n,k}, \mathbb{R}^{m,k}$ after optimally aligning them, where $n$ and $m$ are the respective lengths of $x$ and $x'$ and $k$ is the dimension of the feature space. That is to say, we want to solve the following optimisation problem :
\begin{equation}
DTW(x,x') = \min_{\pi \in A(x,x')} \left(\sum_{(i,j) \in  \pi} d(x_i, x'_j) \right)
\end{equation}
where $\pi$ is an alignment path, $d(x_i, x'_j)$ is the distance between $x_i \text{ and }  x'_j$   and $A(x,x')$ is the set of all possibles paths.

In practice, one can compute DTW by using the dynamic programming solution of the problem as explained by Salvador and al \cite{salvador2007toward}. We first compute the distance matrix $D \in \mathbb{R}^{n,m}$ with:
\begin{multline}
\label{def_DTW}
    D(i,j) = d(x_i-x_j') + \\ min[D(i-1,j-1),D(i-1,j), D(i,j-1)] 
\end{multline}
Then we find the optimal path in D. }

\subsection{Approximation of the DTW}
\label{approxs_sub}
While the advantages of DTW are well-known, its quadratic complexity in both time and space has limited its practical use to small datasets of short time series. To counteract those limitations, some efforts have been made to introduce approximated versions of DTW. \newline
Salvador and al \cite{salvador2007toward} introduced FastDTW, an approximation with linear complexity in both time and space. \mycomment{It involves three main operations : \textbf{coarsening}, which aims at representing the original time series curve with fewer points; \textbf{projection}, consisting in finding a minimum-distance warp path at a lower dimension, using this path to initialize the search of the path in higher dimensions; and \textbf{refinement}, which refines the warp path from low dimension with local modifications.} However, because FastDTW is not differentiable, it cannot be used directly as a loss in gradient based algorithms. \mycomment{ It has also been shown by Wu and al \cite{wu2020fastDTW} to be slower in some real case applications compared to the standard algorithm, named Full or standard DTW.} \newline
To allow for differentiable end-to-end learning with DTW, Cuturi and al \cite{softDTW} introduced SoftDTW. The algorithm computes the soft minimum of all costs spanned by all possible  alignments  between two time series, which leads to a differentiable approximation of the DTW. While the forward pass has a linear time complexity, the backward pass needs to consider all the alignments, resulting in a quadratic complexity in time and a linear complexity in space. The addition of the smoothing factor $\gamma$, also may force more hyperparameter tuning. 

With DTWNet \cite{cai2019DTWnet}, Cai and al introduced a stochastic backpropagation of DTW. They leverage the fact that once the warping path of the DTW is obtained, it is fixed for the iteration, cannot have more than $(n+m)$ elements (if $n$ and $m$ are the respective lengths of the input signals) and is in itself differentiable. While the gradient can be computed in $O(n+m)$, the warping path needs to be obtained, which still requires $O(n.m)$ operations. 

Therefore, while various approaches have been proposed to approximate DTW, to the best of our knowledge none of them enables both differentiability and at most linear complexity in time. 

\subsection{Approximation of distances via neural networks}
\label{approx}

As far as we know, no one has attempted to mimic the DTW with a neural network. However, Courty and al \cite{courty2017learning} similarly approximate the Wasserstein distance using a siamese network to make the squared Euclidean distance of the embedded vectors mimic the Wasserstein distance between the base vectors.
The two input vectors are fed through the same (hence siamese) encoder $\phi$. Then a decoder $\psi$, in that case two fully connected layers, tries to reconstruct the original vector. The encoder learns through the MSE loss, while a KL divergence loss is used for the reconstruction error of the decoder. \mycomment{The overall objective function is as follows: 

\begin{multline}
      \min_{\phi, \psi} \sum_{i} \big\lVert \| \phi(x^1_{i}) - \phi(x^2_{i}) \| ^2 - y_i \big\rVert ^2  \\ +  \lambda \sum_{i} \text{KL}(\psi(\phi(x^1_{i})), x^1_{i}) + \text{KL}(\psi(\phi(x^2_{i})), x^2_{i})
\end{multline}}

Authors choose to use the KL divergence because the Wasserstein distance operates on probability distributions. 
\mycomment{
Courty's and al  method \cite{courty2017learning} has two main advantages over directly predicting the value of the metric with the neural network.} This allows for interpretation of the embedding space, and also fulfills two conditions of a distance (identity and symmetry) since the model is deterministic in inference.

\section{Architecture for learning to approximate DTW}
\begin{figure} 
\centering
\includegraphics[scale = .45]{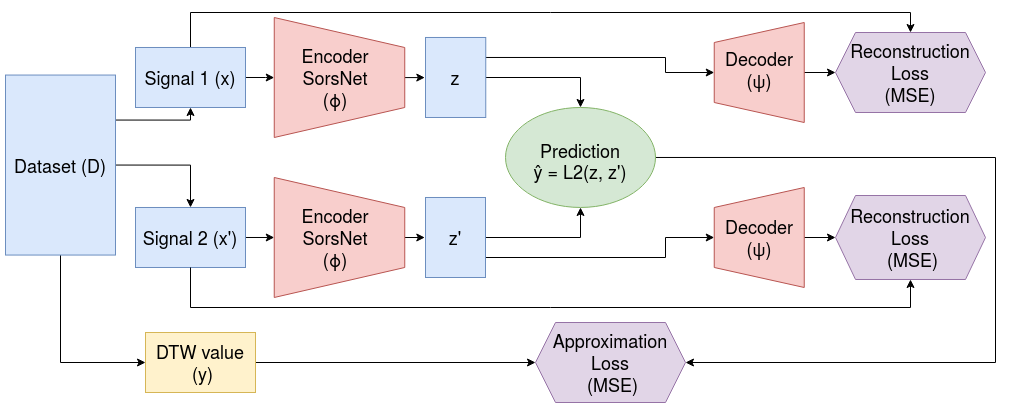}
    \caption{Global architecture of the model. Two signals are drawn from the dataset and encoded by the same encoder. The goal is to get the L2 distance between the encoded vectors as close as possible to DTW between the drawn signals. The encoded signals then pass through a decoder, which tries to reconstruct the original signal. }
    \label{fig:overachi}
\end{figure}

In this section, we introduce the two architectures we use to approximate DTW that will be compared in sections 4 and 5. Because of all the advantages of the method mentioned in \ref{approx}, we first choose to use a similar siamese architecture as in Courty and al \cite{courty2017learning}. Our adapted global architecture is shown in figure \ref{fig:overachi}. 

Contrary to the Wasserstein distance used in Courty and al \cite{courty2017learning}, DTW does not work with probabilities distributions but directly with the time series. Therefore, we use the MSE loss instead of the KL divergence to evaluate the reconstruction error made by the decoder. The goal of the decoder is to force the encoder to keep as much information as possible when projecting the signals. That way, the encoder can not collapse into embedding all the signals to the same representation. It also helps to regularize the training. 
Overall, the encoder $\phi$ takes as input two signals $x \in \mathbb{R}^{L}$ and $x' \in \mathbb{R}^{L'}$ and projects them to two signals $z \in \mathbb{R}^{H}$ and $z' \in \mathbb{R}^{H}$, where $H$ is the hidden dimension. 
Feeding pairs of signals $\{ x_i, x_j' \}_{i,j \in 1,...,n} $ to the model, the global objective function is then, with $ z = \phi(x), z' = \phi(x')$ denoting the encoded signals, $\psi$ the decoder, and $y_{i,j} $ the target DTW value:

\begin{equation}
\min_{\phi, \psi} \underbrace{\sum_{i,j} \big\lVert \| z_i - z_j' \| ^2 - y_{i,j} \big\rVert ^2}_{\textrm{approximation loss}} +\lambda \underbrace{\left( \sum_{i,j} \| \psi(z_i) - x_{i} \| ^2 +  \sum_{i,j} \| \psi(z_j') - x_{j}' \| ^2 \right)}_{\textrm{reconstruction loss}}
\end{equation}
\normalsize
$\lambda$ is a hyperparameter and aims at balancing the losses.

\subsection{Training Procedure}
We describe our training loop with the algorithm \ref{alg:cap}. We directly feed pairs of signals $\{ x_i, x_j' \}_{i, j \in 1,...,n} $ of length $L$ and dimension $d=1$ to the encoder, which processes both signals independently. We use their corresponding DTW distance $ \{y_{i,j} = DTW(x_i, x_j')\} _{i,j \in 1,...,n}$ as label.
Once we have encoded signals, we get the predicted DTW value by taking the Euclidean distance between them and compare it to the reference value via MSE to get the encoder loss. We then use the decoder to get the decoded signals from the encoded ones, then compare them to the input signals to get the decoder loss. We then sum the losses with a balancing parameter $\lambda$ and update the parameters of the encoder and decoder at the same time.

\small
\begin{algorithm}
    \textbf{Input} Mini-batch $\bar{D}$, encoder $\phi$, decoder $\psi$ \\
    \textbf{Output} Updated $\phi$, $\psi$
    \begin{algorithmic}
        \For { $ \forall \: (x, \: x')$ in $\bar{D}$}:
        \State $y \gets DTW(x, x') $ \Comment{Reference value, pre-calculated in practice to save time}
        \State $z, \: z' \gets \phi(x, \: x') $ \Comment{$ z \in \mathbb{R}^H$}
        \State $y_{pred} \gets \| z - z' \| _{2}$
        \State $L_{encoder} \gets MSE(y_{pred} , y)$
        \State $\hat{x}, \: \hat{x}' \gets \psi(z, \: z') $
        \State $L_{decoder} \gets MSE(\hat{x}, x) + MSE(\hat{x}', x')$
        \State $L \gets L_{encoder} + \lambda * L_{decoder}$
        \State update $\phi$, $\psi$ with $L$
        \EndFor
    \end{algorithmic}
    \caption{Training loop for each mini-batch for the approximation of DTW.}
    \label{alg:cap}
\end{algorithm}
\normalsize

\subsection{Encoder architecture}
\label{subsec:encoder}
The global architecture of our approach is independent of the type of time series we train it on. On the other hand, if we want a reliable approximation, the encoder needs to be able to project the signals meaningfully, and therefore must be adapted to each type of data. In our case, we choose to focus on EEG data. As a result, we use SorsNet introduced by Sors and al \cite{SorsNet} as encoder. SorsNet is a 1D convolutional neural network consisting of a series of blocks. Each block contains a convolutional layer, a batch normalization layer and a ReLU activation function. We choose SorsNet because it has been shown to work well on sleep staging on EEG signals (\cite{SorsNet}), thus we assume that the architecture allows for a good representation of EEG data. The network is also fully convolutional, with kernel sizes, strides and padding carefully chosen to always get a projected vector $ z \in \mathbb{R}^{1,H}$ as long as the length of the time series is less or equal than 3000, the usual size for EEG data. This permits us to use the same model with the same weights for time series of different lengths, thus allowing to mimic the DTW ability to compare time series of different lengths. Finally, the network being fully convolutional also enables low inference time. 

\subsection{Decoder}
\label{dec_sec}
We want to force the encoder to learn a meaningful embedding that keeps as much information about the original signals as possible, in order to improve the accuracy of the approximation of DTW. \newline 
Inspired by Thill and al \cite{thill2021temporal}, we first use an upsampling layer so that $z$ is of the same dimension as the input signal $x$, then use a Temporal Convolutional Network (TCN, Bai and al \cite{bai2018empirical}) to decode $z$ and try to retrieve $x$. 
We use $ q = [32, 16, 8, 4, 2, 1] $ as dilation rates and $k = 20 $ as kernel size. The choice of a TCN allows our decoder to be independent of the length of time series $x$ since all the layers are convolutional. 

\subsection{Direct Regression}
While the architecture of our encoder-decoder allows comparing signals of different lengths, a simpler architecture may work better on signals of fixed lengths. To investigate this, we also introduce a simpler architecture that we call \textbf{direct regression}. It takes as input pairs of signals $\{ x_i, x_j' \}_{i,j \in 1,...,n} $, concatenates them to get a tensor $ x_{cat} \in  \mathbb{R}^{B, L, 2}$ with $B$ the batch size, then feeds $x_{cat}$ as input to the SorsNet encoder. Afterwards, a dense layer with batch normalization and ReLU activation processes the tensor, before a final dense layer outputs the predicted DTW value. In this case, we do not need a decoder since we directly get the value to predict. Everything else is kept identical to the siamese encoder-decoder architecture.

\section{Experimental Setup}

\subsection{Datasets}
\label{SleepEDF_DS}
While ideally our model should be able to approximate the DTW no matter the origin of the time series, we decide to first focus only on sleep data. We choose to use the SleepEDF-78 dataset \cite{SleepEDF}, which contains recording of various sensors during sleep. The participants were involved in two studies: Sleep Cassette, which focuses on age effect on sleep and  Sleep  Telemetry, which focuses on the impact of temazepam on  sleep. For  these  two  datasets,  each  PSG  file  contains  two  EEG channels (Fpz-Cz, Pz-Oz) sampled at 100 Hz, one EOG channel and one chin EMG channel. We decided, following the literature (Phan and al \cite{jointcnn}, Tsinalis and al \cite{tsinalis2016automatic}) to use only the Sleep Cassette files. To train our model, we are however able to use all the different channels, while previous studies focus on the Fpz-Cz channel only. We randomly split the patients, keeping 70 patients for training and validation and 8 patients for testing. For each patient, we split the signals in cuts of size $L$ along a randomly chosen signal. The dataset is then made of $N$ randomly selected cuts.

\textbf{Data preprocessing} \newline
\label{preproc_sub}
Since we use multiple channels of the sleep files, we have to process various types of data. This creates scaling problems, which is to say that some series will have values in much bigger ranges than others. Moreover, it will lead to big ranges for the reference DTW values and thus for the training loss. To face this difficulty, we preprocess the dataset as follows. We first clip all the values in the dataset $Ds$ i.e., for every value $a_k$ in each signal $x_i$ in $Ds$, we compute $p_1$ and $p_{99}$, respectively 1st and 99th percentile. \newline Then \( \forall a_k \in Ds:\)
\small
\begin{equation}
    a_k = 
    \begin{cases}
      p_1 & \text{if} \ a_k < p_1 \\
      p_{99} & \text{if} \ a_k > p_{99} \\
      a_k & \text{otherwise}
    \end{cases}
  \end{equation}
 \normalsize
 This allows to limit the impact of outliers without losing much information. We then apply the min-max normalization. It projects all the data to \( [0,1] \), which bounds the reference DTW to \( [0, L] \) where $L$ is the length of the time-series $x$ since DTW is ultimately the sum of Euclidean distances along the alignment path. Knowing the lower and upper bounds of the DTW allows us to normalize its values to \( [0,1] \), greatly helping to balance the training losses. 
\textbf{Creation of the DTW matrix} \newline
We then fill the ground truth DTW matrix $Y_{DTW}$. To do so, we randomly choose $N_{pairs}$ pairs of signals and fill the corresponding parts of the ground truth matrix with DTW results between the pairs of selected signals, following Courty and al \cite{courty2017learning}. 

\subsection{Training Parameters}
\label{train_det_sec}
To speed up the training, we restrain the signals to length $L = 1000$ by randomly slicing the signal. We select $N = 10,000$ signals as explained in section \ref{SleepEDF_DS} for the train set and randomly select $N_{pairs} = 10^6$ over the $10^8$ possible pairs in order to train the model on a decent number of signals without overfitting. It also limits the training time. We do the same for the validation set and the test set with 1000 signals and 100,000 pairs. \newline
We use SorsNet as encoder, setting the dropout to 0 and replacing the classification layer by a dense layer with $H = 500$. We define the decoder as described in section \ref{dec_sec} and use Adam optimizer with a $10^{-5}$ learning rate to optimize both the encoder and the decoder parameters at the same time. We set the batch size to 128 and $\lambda$ to 1. We train for 50 epochs with an early stopping if the validation loss does not improve for 8 straight epochs. 
\newline
Note that because of time constraints (training lasts for approximately 20 hours on one TITAN V), no extensive hyperparameter search was done. 

\section{Study of performance on downstream tasks}
\subsection{Illustration of Efficiency and Approximation \label{NN_sec}
Properties on a Nearest Neighbour Retrieval Task}
In itself, the output value of DTW is not our main goal. What matters is the ability to compare time series and rank similarity between time series. Therefore, instead of comparing the raw values of our approximation with DTW, once we have trained our model we study its performance on downstream tasks.

We want to study how our model can compare series and how close it is to DTW. To do so, we first select $N_t$ signals in the test set. We then fit a nearest neighbor algorithm  \footnote{https://scikit-learn.org/stable/modules/neighbors.html} using as custom metric our model (DeepDTW) on the test set. We do the same operation using DTW (we use the implementation from pyts by Faouzi and al \cite{pyts}) as custom metric. 
We select the top1 nearest neighbor $ \Tilde{x}_{top1} $ for all signals \(x \in N_t\) with \( x \neq \Tilde{x}_{top1} \) and evaluate the number of times 
\( \Tilde{x}_{top1} \) is in the top 5 ranking of nearest neighbors according to DTW. Since we use random subsets of the test set, we run the same experiment 8 times for increasing numbers of signals $N_t$. We also add the two main approximations of DTW, FastDTW and SoftDTW, to the comparison. Since SoftDTW running time is dependent on the hyperparameter $\gamma$ (the closer $\gamma$  to 0, the more faithful SoftDTW to the standard DTW, but also the slower as explained by Cuturi and al \cite{softDTW}), we choose a middle ground with $\gamma = 0.1$.

One of the major perks of DTW is its ability to compare time series of different lengths, and so a good approximation should mimic this feature. 
To study this property, we modify our dataset by restricting the size of EEG signals to random lengths following the uniform sampling method from Tan and al\cite{var_tsc} i.e., $ \forall x \in N_t$ , $ x \in \mathbb{R}^L$ with $L \in [500 ; 1000]$.
While in inference no change is required for our siamese architecture to compute signals of varying lengths, we have to pad the signals to the size of the longest in a batch to make the direct regression architecture work.
We show the results in table \ref{tab:tsc_var_len}. 
We can see that the direct regression (DeepDTW Direct) model learns to order signals the closest to DTW, even outperforming FastDTW when the task is easy or moderately easy (50-200 signals). FastDTW is less impacted by the task getting harder and is the best with 400 and 600 signals in the set. Both our approximations outperform SoftDTW no matter the number of signals, with the direct approach above by $\sim$43 percentage points (pp) when the task is easy  and still  $\sim$21pp above in the hardest setting, while the siamese approach is $\sim$20pp above at 50 signals and about equal at 600 signals.

It illustrates how our model can mimic DTW ability to compare series of different lengths well enough.

\begin{table*}
	\begin{center}
		\begin{tabular}{|c|c|c|c|c|}
		\hline
			N signals & SoftDTW(0.1) & DeepDTW Direct & DeepDTW Siamese & FastDTW \\
			\hline
			50 & 42.25 ± 4.79 & \textbf{85.75 ± 4.74} & 62.0 ± 6.48 & 74.25 ± 7.45 \\
			100 & 28.12 ± 3.82 & \textbf{75.25 ± 1.39} & 45.25 ± 6.76 & 65.25 ± 3.67 \\
			200 & 23.06 ± 2.16 & \textbf{61.0 ± 2.76} & 33.0 ± 3.22 & 57.56 ± 4.44 \\
			400 & 20.22 ± 1.33 & 47.31 ± 3.12 & 22.84 ± 2.43 & \textbf{52.47 ± 1.83} \\
			600 & 18.0 ± 1.43 & 39.88 ± 2.33 & 18.52 ± 0.83 & \textbf{49.75 ± 1.86} \\
			\hline
		\end{tabular}
	\end{center}
	\caption{Percentage of time the closest different signal of length $ L \in [500;1000]$ of a given signal according to our model is among the top 5 closest according to DTW. N signals is the number of signals in the test set.}
	\label{tab:tsc_var_len}
\end{table*}

\subsection{Sleep Staging}

To complete the study of the faithfulness of our approximation, we evaluate how our model can be used in time-series classification context. We choose the sleep staging task to do so. It consists in classifying segments of sleep data in different classes. 

\textbf{Dataset} \newline
\label{SleepEDF_section}
We use the test set from the SleepEDF dataset, with the same processing as in section \ref{SleepEDF_DS}. This time we use the labels of the states of sleep. The classes in SleepEDF include wake (W), rapid eye movement (REM), four stages of different deep sleep (N1, N2, N3, N4), M (movement time) and '?' (no data). Following Mousavi and al \cite{mousavi2019sleepeegnet}, we merge the stages N3 and N4, and remove the sequences labelled as M and '?'. It results in sequences of signals of \( length = 3000 \) distributed in 5 different classes.

\textbf{KNN classification} \newline
\label{KNN_clf_sedf}
We want to compare our approximation to DTW and its other main approximations for sleep staging. To do that, we create 4 instances of KNN classifiers, each with a different base metric: standard DTW, our model with the direct architecture (DeepDTW Direct), our model with the siamese architecture (DeepDTW Siamese) and FastDTW. We then run 5 iterations of the following experiment: select a number $N$ of random signals in the set and split them in training and test sets at 50/50 proportion, fit the KNN instance of each metric on the training set and compute the corresponding macro F1 score (MF1) on the test set. 
We modify the SleepEDF test set to get time series of varying length in the same way as in section \ref{NN_sec}. 
We show the results on the SleepEDF dataset in table \ref{tab:tsc_dir_siam}.

\mycomment{
\begin{figure}[htbp] 
\centering
    \includegraphics[scale = .348]{figures/Sedf_classif_F1_macro_DTW_classic_nsig_full_NEW.png}
    \caption{Macro F1 score of sleep staging on SleepEDF with KNN using various metrics as base for the KNN. We run 5 iterations. For any given iteration, all the methods use the same data.  }
    \label{fig:sledf_nsig}
\end{figure}}

\begin{table*}
	\begin{center}
		\begin{tabular}{|c|c|c|c|c|}
			\hline
			N signals & DTW & DeepDTW Direct & DeepDTW Siamese & FastDTW \\
			\hline
			500 & 0.25 ± 0.02 & 0.26 ± 0.04 & 0.21 ± 0.02 & 0.26 ± 0.01 \\
			1000 & 0.34 ± 0.02 & 0.33 ± 0.01 & 0.37 ± 0.02 & 0.35 ± 0.02 \\
			2000 & 0.44 ± 0.01 & 0.44 ± 0.01 & 0.43 ± 0.01 & 0.43 ± 0.01 \\
			\hline
		\end{tabular}
	\end{center}
	\caption{Macro F1 score of sleep staging on SleepEDF with KNN using various metrics as base for the KNN. We run 5 iterations. For any given iteration, all the methods use the same data.}
	\label{tab:tsc_dir_siam}
\end{table*}

Overall, both our approximation and FastDTW behave very similarly to standard DTW. Small variations due to variance aside, we can see that both our models and FastDTW lead to very similar classification score to DTW, which shows that they tend to compare series in the same way. 



\subsection{Computation Time Study}
\begin{figure} 
\centering
    \includegraphics[scale = .5]{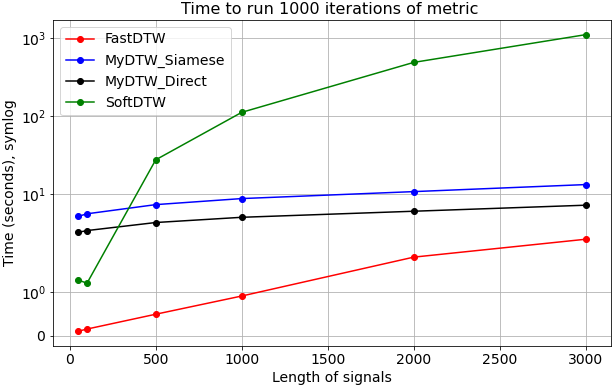}
    \caption{Time needed in seconds to run 1000 computations \textbf{on CPU} of the metric, depending on the lengths of uni-dimensional signals.}
    \label{fig:time_len}
\end{figure}

While the main drawback of FastDTW is the fact that it is not differentiable, the main drawback of SoftDTW is its low computation speed. To illustrate this point, we study the time needed to compute DTW between two uni-dimensional time series 1000 times. To fairly compare all the metrics, all the experiments are run only on CPU. We keep $\gamma = 0.1$ for SoftDTW. We show the results in figure \ref{fig:time_len}. We can see how the computation times needed for both versions of our models (in blue and black) are very slowly increasing with the size of the time series, while FastDTW and especially SoftDTW running times are quickly increasing. At \(L = 3000 \), the standard size for sleep staging, our direct model is 100 times faster than SoftDTW. Our direct model is only approximately 3 times slower than FastDTW, can be run on GPU, and is differentiable, as shown in the following section.

\subsection{Differentiability} 
\label{diff_sec}
The goal of our model is to be accurate, fast, and differentiable.
In this section, we illustrate the latter. 

\begin{figure} 
\centering
    \includegraphics[scale = .5]{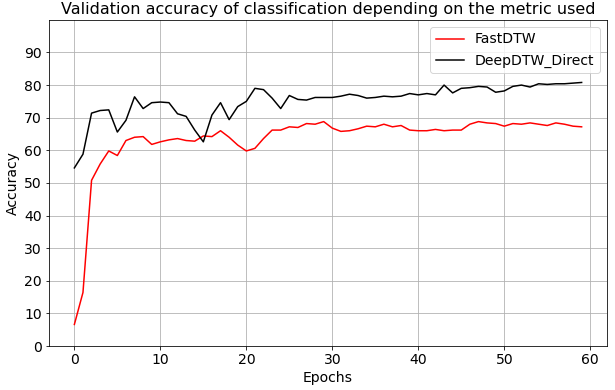}
    \caption{Accuracy score of prototype-based classification on the validation set during the training of class specific prototypes. The metric is used to compute the distance between an input signal and prototypes, and so is crucial for the training. FastDTW is used to compute the alignment path following Cai and al \cite{cai2019DTWnet} (see section \ref{diff_sec}).}
    \label{fig:TSC_learnig}
\end{figure}

Chang and al \cite{chang2021learning} introduced a way to learn class specific prototypes in order to classify time series. For a given dataset $D$, they learn as many prototypes $p$ as the number of classes $k$ in the dataset: the inter-class distance between prototypes should be as large as possible, while at the same time a prototype should represent its class well enough to get good classification results. Once prototypes are learned, times series can be classified by using the nearest neighbor algorithm with the prototypes. They are learned by computing DTW between a given signal $x_n \in D$ and each prototype $p_k$ corresponding to each class $k$. Since the idea is to learn the prototype end-to-end, to circumvent the non-differentiabilty of DTW, the authors choose to differentiate DTW by using the determined form along the warping path, i.e., the sum of Euclidean distances of paired points as done by Cai and al \cite{cai2019DTWnet}.  
\newline \mycomment{Since authors give access to their code, we have applied their method to the SleepEDF dataset, and changed DTW computation from differentiation along the warping path to approximation with our model and backpropagate the gradient through our model.}
We apply the method to the SleepEDF dataset and compare the classification score obtained by computing DTW with our approximation. We show the results in figure \ref{fig:TSC_learnig}. We can see that with our model used as distance to compare the signals, we learn prototypes that represent the classes better since the classification accuracy is higher. It is also faster to train, even on CPU (854 minutes for our method, versus 1773 minutes). 

We have shown that our approximation performs well in time series classifications tasks, is fast, and can be used to learn end-to-end. However, a good approximation of DTW should perform well independently of the dataset. This is what we illustrate in the following section. 

\subsection{Adaptation to others datasets}
In this section, we study the generalization capacity of our model to similar EEG data.

\textbf{Dataset} \newline
Following other sleep staging related contributions (Olesen and al \cite{olesen2021automatic}, Phan and al \cite{phan2021xsleepnet}, Eldele and al \cite{eldele2021attention}) we evaluate the generalization capabilities of our model to another widely used EEG dataset: SHHS, introduced by Quan and al \cite{quan1997sleep} and later updated by Zhang and al \cite{zhang2018national}. 
SHHS is a multichannel health signal database, aimed at studying the effect of sleep-disordered breathing on cardiovascular diseases. There are two rounds of PSG records in the dataset, SHHS-1 for Visit 1 and SHHS-2 for Visit 2. We only focus on the first set in this section. It contains 5791 subjects. 
Similar to other databases like SleepEDF annotated with the R\&K rule, N3 and N4 stages were merged into N3 stage and MOVEMENT (M) and UNKNOWN (?) signals were removed. As we did in section \ref{SleepEDF_section}, for each signal we randomly choose a channel among the pre-selected ones (EEG, EOG(L), EOG(R), ECG and EMG channels for this study).

\textbf{Nearest Neighbor Retrieval} \newline
We reproduce the experiments from section \ref{NN_sec}, this time on the SHHS dataset. We choose to only use the direct architecture as it gave the best results on SleepEDF. To study how our model transfers its knowledge to other datasets, we first use the best model learned on SleepEDF according to the nearest neighbor test, and directly use it without fine-tuning to do the same test on the SHHS dataset.
We use the same preprocessing as in section \ref{preproc_sub}.



We also learn our approximation model on the SHHS dataset in the same way as in section \ref{train_det_sec} and do the nearest neighbor experiment on a separated test set. 
We summarize the results in table \ref{tab:nn_test}. The model learned on SleepEDF and tested on SHHS gives almost identical results to the one learned on SHHS, showing that for similar data with consistent preprocessing, our approximation model generalizes very well to new data. 

\begin{table*}[ht]
\centering
    \begin{tabular}{|c|c| c |c |c} 
     \hline
     N signals &  Model trained on SHHS  &  Model trained on SleepEDF  \\ [0.5ex] 
     \hline
     {50} & {$0.89 \pm 0.04$}    & {$0.86 \pm 0.03$}    \\ 
     {100} & {$0.75 \pm 0.05$}     & {$0.74 \pm 0.04$}   \\
     {200} & {$0.61 \pm 0.03$}     & {$0.62 \pm 0.02$}   \\
     {400} & {$0.50 \pm 0.03$}      & {$0.48 \pm 0.02$}    \\
     {600} & {$0.44 \pm 0.02$}     & {$0.43 \pm 0.02$}   \\ 
      \hline
    \end{tabular}
\caption{Nearest neighbor test score on SHHS. The model trained of SleepEDF is not fine-tuned.}
\label{tab:nn_test}
\end{table*}
\textbf{KNN Classification on SHHS} \newline
We reproduce the experiment from section \ref{KNN_clf_sedf} on the SHHS dataset. We apply exactly the same processing on SHHS, generating time series of varying length. We compare the classification performance of 3 KNNs, one based on FastDTW, one on our direct regression model learned on SleepEDF (DeepDTW SleepEDF) and one learned on SHHS (DeepDTW SHHS).
We show the results in table \ref{tab:tsc_shh}. Our models are very close to each other, showing that they also generalize well in the classification context. 

\mycomment{
\begin{figure}[htbp] 
    \includegraphics[scale = .348]{figures/SHHS_classif_F1_macro_DTW_classic_nsig_full_NEW.png}
    \caption{Macro F1 score of sleep staging on SHHS with KNN using various metrics as base for the KNN. DeepDTW\_Direct\_SleepEDF stands for the model learned on the SleepEDF set and not fine-tuned, while DeepDTW\_Direct\_SHHS indicates the model with direct architecture trained on SHHS.}
    \label{fig:shhs_nsig}
\end{figure}
}

\begin{table*}[ht]
	\begin{center}
		\begin{tabular}{|c|c|c|c|}
		    \hline
			N signals & DeepDTW SHHS & DeepDTW SleepEDF & FastDTW \\
			\hline
			500 & 0.250 ± 0.02 & 0.246 ± 0.02 & 0.249 ± 0.01 \\
			1000 & 0.329 ± 0.01 & 0.327 ± 0.01 & 0.342 ± 0.04 \\
			2000 & 0.427 ± 0.01 & 0.437 ± 0.01 & 0.472 ± 0.01 \\
			\hline
		\end{tabular}
	\end{center}
	\caption{Macro F1 score of sleep staging on SHHS with KNN using various metrics as base for the KNN. DeepDTW SleepEDF stands for the model learned on the SleepEDF set and not fine-tuned, while DeepDTW SHHS indicates the model trained on SHHS. Both use direct architecture.}
	\label{tab:tsc_shh}
\end{table*}

\section{Conclusion and future works} 
In this paper, we presented and compared two architectures to approximate DTW. The first one is done by creating an embedding in which the Euclidean distance mimics the DTW. Such an embedding is obtained by training a siamese encoder-decoder model to both regress the DTW value and retrieve the original signals from the embedded ones. The second method concatenates signals to directly predict DTW value, allowing for better retrieval performance and slightly faster training time, and is therefore a better approach.  
We showed how our approximations can be used in an end-to-end training, are faster and more faithful to DTW than other approximations, and perform well in time series classification tasks. Finally, we also showed that we can extend the results to similar datasets. 
\newline
However, although multidimensional time series are quite common in DTW use cases, they were not addressed in this paper and are left for future works. Particularly, being able to embed signals in a space where the Euclidean distance mimics DTW no matter the length or number of dimensions of the signals would be the end goal. 
\newline 
Finally, a perfect approximation of DTW should be usable on various types of data without the need for fine-tuning. Like it has been done in the literature for natural language processing, we leave for future work to build a huge dataset of various types of time series and create a generalist model able to approximate DTW on all those types of data.  
\bibliography{bib_}
\bibliographystyle{unsrt}

\end{document}